\documentclass[journal]{IEEEtran}

\usepackage[utf8]{inputenc}
\usepackage{graphicx}
\usepackage{amsmath}
\usepackage{algorithmic}
\usepackage{array}
\usepackage[caption=false,font=footnotesize]{subfig}
\usepackage{stfloats}
\usepackage{multirow}
\usepackage{url}
\usepackage{cite}

\makeatletter
\newcommand*\titleheader[1]{\gdef\@titleheader{#1}}
\AtBeginDocument{
\let\st@red@title\@title
\def\@title{
\bgroup\normalfont\footnotesize\centering\@titleheader\par\egroup
\vskip0.5em\st@red@title}
}
\makeatother

\title{Self-Supervised Vision-Based Detection of the Active Speaker as Support for Socially-Aware Language Acquisition}
\titleheader{This article has been accepted for publication in a future issue of this journal, but has not been fully edited. Content may change prior to final publication. Citation information: DOI 10.1109/TCDS.2019.2927941, IEEE Transactions on Cognitive and Developmental Systems}

\author{Kalin Stefanov, Jonas Beskow and Giampiero Salvi
\thanks{This research was supported by the CHIST-ERA project IGLU and KTH SRA ICT The Next Generation. We would like to acknowledge the NVIDIA Corporation for donating the GeForce GTX TITAN cards used for this research, and the Swedish National Infrastructure for Computing (SNIC) at the Parallel Data Center (PDC) at KTH for computational time allocation. We would like to thank the anonymous reviewers for their insightful comments.}
\thanks{K. Stefanov is with the Institute for Creative Technologies, University of Southern California, Los Angeles, USA. J. Beskow is with the department of Speech, Music and Hearing, KTH Royal Institute of Technology, Stockholm, Sweden. G. Salvi is with the department of Electronic Systems, NTNU Norwegian University of Science and Technology, Trondheim, Norway.}
\thanks{E-mail: kalins@kth.se, beskow@kth.se and giampi@kth.se}}

\IEEEpubid{\begin{minipage}{\textwidth}\ \\\\
\footnotesize\centering 2379-8920 (c) 2019 IEEE. Personal use is permitted, but republication/redistribution requires IEEE permission. See http://www.ieee.org/publications\_standards/publications/rights/index.html for more information.
\end{minipage}}

\begin{document}

\maketitle

\begin{abstract}
This paper presents a self-supervised method for visual detection of the active speaker in a multi-person spoken interaction scenario. Active speaker detection is a fundamental prerequisite for any artificial cognitive system attempting to acquire language in social settings. The proposed method is intended to complement the acoustic detection of the active speaker, thus improving the system robustness in noisy conditions. The method can detect an arbitrary number of possibly overlapping active speakers based exclusively on visual information about their face. Furthermore, the method does not rely on external annotations, thus complying with cognitive development. Instead, the method uses information from the auditory modality to support learning in the visual domain. This paper reports an extensive evaluation of the proposed method using a large multi-person face-to-face interaction dataset. The results show good performance in a speaker dependent setting. However, in a speaker independent setting the proposed method yields a significantly lower performance. We believe that the proposed method represents an essential component of any artificial cognitive system or robotic platform engaging in social interactions.
\end{abstract}

\begin{IEEEkeywords}
active speaker detection and localization, language acquisition through development, transfer learning, cognitive systems and development
\end{IEEEkeywords}

\section{Introduction}
\label{sec:introduction}
\IEEEpubidadjcol
\IEEEPARstart{T}{he} ability to acquire and use language in a similar manner as humans may provide artificial cognitive systems with a unique communication capability and the means for referencing to objects, events and relationships. In turn, an artificial cognitive system with this capability will be able to engage in natural and effective interactions with humans. Furthermore, developing such systems can help us further understand the underlying processes in language acquisition during the initial stages of the human life. As mentioned in~\cite{bosch09}, modeling language acquisition is very complex and should integrate different aspects of signal processing, statistical learning, visual processing, pattern discovery, and memory access and organization.

According to many studies (e.g.,~\cite{steels00}) there are two alternatives to human language acquisition --- individualistic learning and social learning. In the case of individualistic learning, the infant exploits the statistical regularities in the multi-modal sensory inputs to discover linguistic units such as phonemes and words and word-referent mappings. In the case of social learning, the infant can determine the intentions of others by exploiting different social cues. Therefore, in social learning, the participants in the interaction with the infant play a crucial role by constraining the interaction and providing feedback.

From a social learning perspective, the main prerequisite for language acquisition is the ability to engage in social interactions. For an artificial cognitive system to address this challenge, it must at least 1) be aware of the people in the environment, 2) detect their state: \emph{speaking} or \emph{not speaking}, and 3) infer possible objects the active speaker is focusing attention on.

\IEEEpubidadjcol
In this study we address the problem of detecting the active speaker in a multi-person language learning scenario. The auditory modality is fundamental for this task and much research has been devoted to audio-based active speaker detection (Section~\ref{subsec:active_speaker_detection}). In this study, however, we propose to take advantage of the temporal synchronization of the visual and auditory modalities in order to improve the robustness of audio-based active speaker detection. The paper proposes and evaluates three \emph{self-supervised} methods that use the auditory input as reference in order to learn an active speaker detector based on the visual input alone. The goal is not to replace the auditory modality, but to complement it with visual information whenever the auditory input is unreliable.

In order to impose as little constraints as possible on the social interaction, we have two requirements for the proposed methods. The first is that any particular method must operate in real-time (possibly with a short lag), which in practice means that the method should not require any future information. The second requirement is that the methods should make as few assumptions as possible about the environment in which the artificial cognitive system will engage in social interactions. Therefore, the methods should not assume noise-free environment, known number of participants in the interaction, or known spatial configuration. The proposed methods address the requirements for engagement in social interactions outlined above, by detecting the people in the environment and detecting their state --- speaking or not speaking. In turn, this information is a prerequisite to hypothesizing the possible objects a speaking person is focusing his/her attention on, which has been shown to play an important role in language acquisition (Section~\ref{subsec:language_acquisition}).

The rest of the paper is organized as follows. First we examine previous research that forms the context for the current study in Section~\ref{sec:related_work}, then we describe the proposed methods in Section~\ref{sec:methods}. The experiments we conducted are described in Section~\ref{sec:experiments} and the results of these experiments are presented in Section~\ref{sec:results}. Discussion on the used evaluation metric, together with the assumptions made can be found in Section~\ref{sec:discussion}. We conclude the paper in Section~\ref{sec:conclusions}.

\section{Related Work}
\label{sec:related_work}
This section is divided in two parts. First we introduce research on language acquisition which supports our motivation to build an active speaker detector for a language learning artificial cognitive system. In the second part of the section we turn our focus on research related to the problem of identifying the active speaker through visual and auditory perceptual inputs.

\subsection{Language Acquisition}
\label{subsec:language_acquisition}
The literature on language acquisition offers several theories of how infants learn their first words. One of the main problems which researchers face in this field is the \emph{referential ambiguity} as discussed for example in~\cite{clerkin16,pereira14,yurovsky13}. Referential ambiguity stems from the idea that infants must acquire language by linking heard words with perceived visual scenes, in order to form word-referent mappings. In everyday life however, these visual scenes are highly cluttered which results in many possible referents for any heard word, within any learning event~\cite{quine13,bloom00}. Similarly, many computational models of language acquisition are rooted in finding statistical associations between verbal descriptions and the visual scene~\cite{roy02,yu04,clerkin16,rasanen15}, or in more interactive robotic manipulation experiments~\cite{salvi12}. However, nearly all of them assume a clutter-free visual scene, where objects are observed in isolation on a simplified background (often white table).

Different theories offer alternative mechanisms through which infants reduce the uncertainty present in the learning environment. One such mechanism is statistical aggregation of word-referent co-occurrences across learning events. The problem of referential ambiguity within a single learning event has been addressed by Smith et al.~\cite{smith08,smith14}, suggesting that infants can keep track of co-occurring words and potential referents across learning events and use this aggregated information to statistically determine the most likely word-referent mapping. However, the authors argued that this type of statistical learning may be beyond the abilities of infants when considering highly cluttered visual scenes. In order to study the visual scene clutter from the infants' perspective, Pereira et al.~\cite{pereira14} and Yurovsky et al.~\cite{yurovsky13} performed experiments in which the infants were equipped with a head-mounted eye-tracker. The conclusion was that some learning events are not ambiguous because there was only one dominant object when considering the infants' point of view. As a consequence, the researchers argued that the input to language learning must be understood from the infants' perspective, and only regularities that make contact with the infants' sensory system can affect their language learning. Although not related to language acquisition, an attempt at modeling the saliency of multi-modal stimuli from the learner's (robot's) perspective was proposed in~\cite{ruesch08}. This bottom up approach is based exclusively on the statistical properties of the sensory inputs.

Another mechanism to cope with the uncertainty in the learning environment might be related to social cues to the caregivers' intent, as mentioned in the above studies. Although a word is heard in the context of many objects, infants may not treat the objects as equally likely referents. Instead, infants can use social cues to rule out contenders to the named object. Yu and Smith~\cite{yu16} used eye-tracking to record gaze data from both caregivers and infants and found that when the caregiver visually attended to the object to which infants' attention was directed, infants extended the duration of their visual attention to that object, thus increasing the probability for successful word-referent mapping.

Infants do not learn only from interactions they are directly involved in, but also observe and attend to interactions between their caregivers. Handl et al.~\cite{handl13} and Meng et al.~\cite{meng17} performed studies to examine how the body orientation can influence the infants' gaze shifts. These studies were inspired by large body of research on gaze following which suggests that infants' use others' gaze to guide their own attention, that infants pay attention to conversations, and that joint attention has an effect on early learning. The main conclusion was that static body orientation alone can function as a cue for infants' observations and guides their attention. Barton and Tomasello~\cite{barton91} also reasoned that multi-person context is important in language acquisition. In their triadic experiments, joint attention was an important factor facilitating infants' participation in conversations; infants were more likely to take a turn when they shared a joint attentional focus with the speaker. Yu and Ballard~\cite{yu04} also proposed that speakers' eye movements and head movements among others, can reveal their referential intentions in verbal utterances, which could play a significant role in an automatic language acquisition system.

The above studies do not consider how infants might know which caregiver is actively speaking and therefore requires attention. We believe that this is an important prerequisite to modeling automatic language acquisition. The focus of the study described in this paper is, therefore, to investigate different methods for inferring the active speaker. We are interested in methods that are plausible from a developmental cognitive system perspective. One of the main implications is that the methods should not require manual annotations.

\begin{figure}[!t]
\centering
\includegraphics[width=\columnwidth]{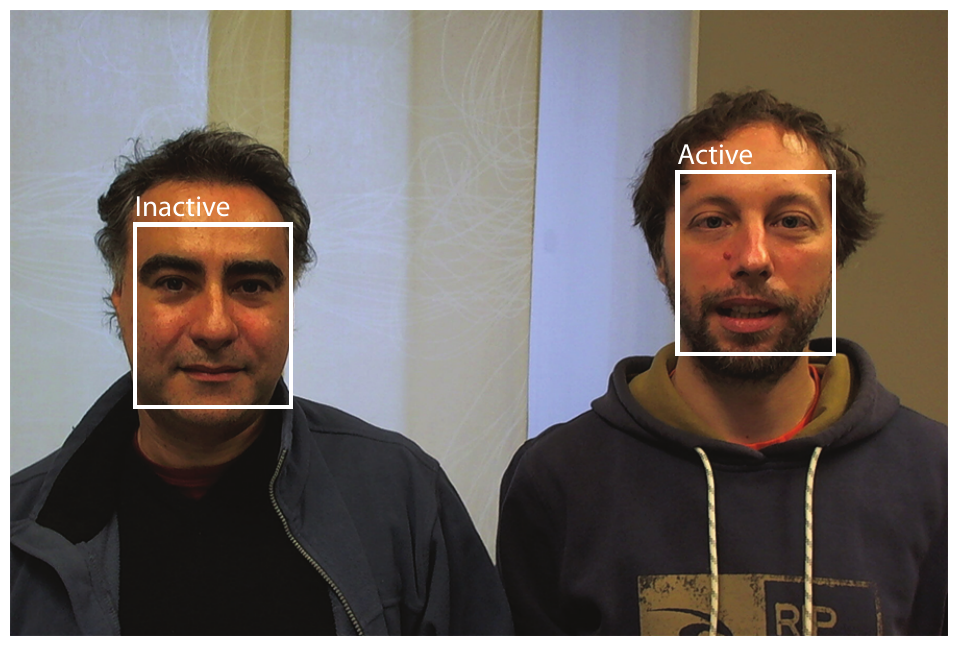}
\caption{Example of an output of a visual active speaker detector.}
\label{fig:detection_example}
\end{figure}

\begin{figure*}[!t]
\centering
\includegraphics[width=\textwidth]{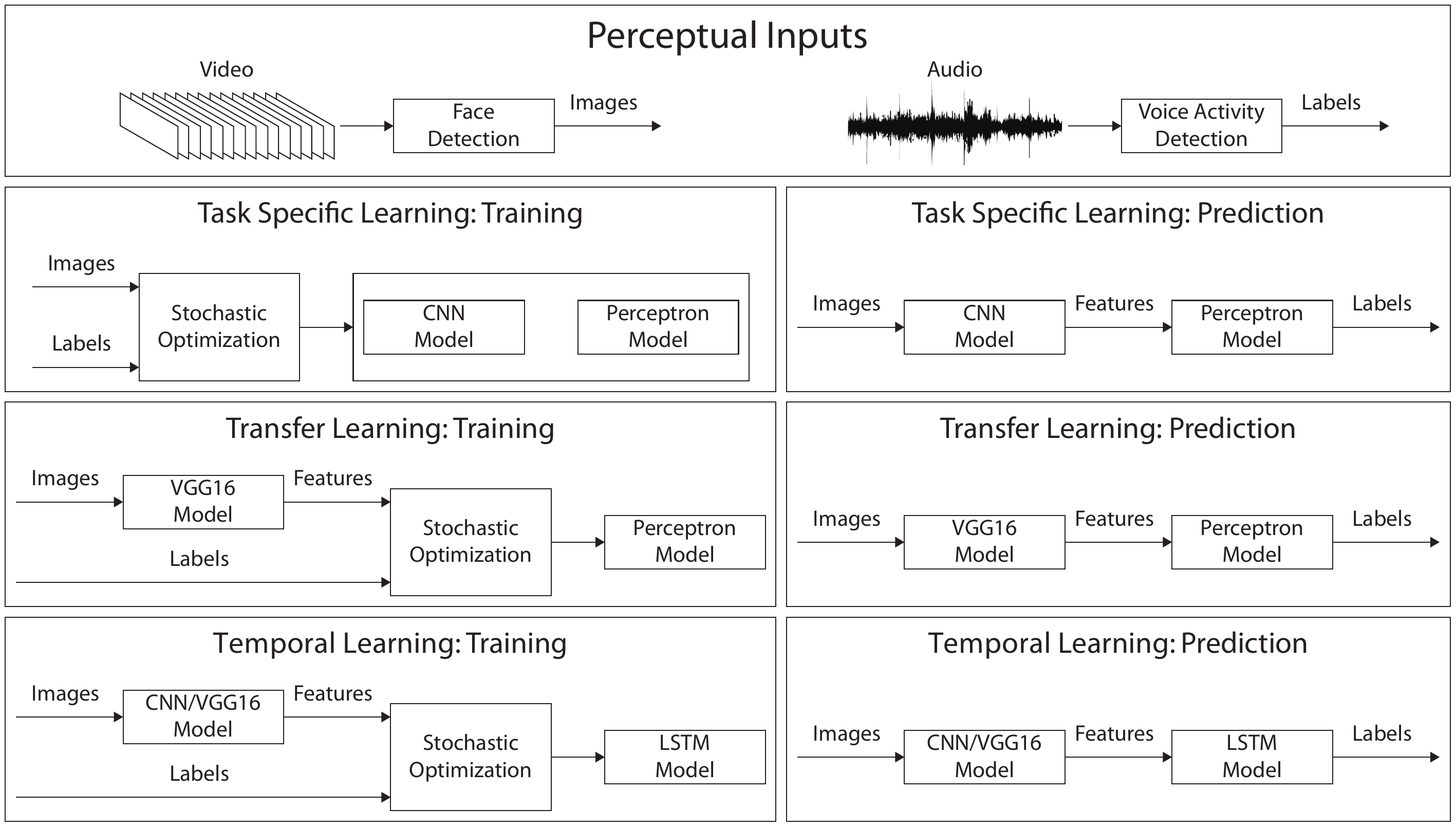}
\caption{Approaches to visual active speaker detection considered in the study. In the first row are the perceptual inputs automatically extracted from the video and audio streams. These inputs are passed to the task specific learning (second row), transfer learning (third row) and temporal learning (forth row) methods.}
\label{fig:methods}
\end{figure*}

\subsection{Active Speaker Detection}
\label{subsec:active_speaker_detection}
Identifying the active speaker is important for many applications. In each area, different constraints are imposed to the methods. Generally, there are three different approaches: audio-only, audio-visual, and approaches that use other forms of inputs for detection.

Audio-only active speaker detection is the process of finding segments in the input audio signal associated with different speakers. This type of detection is known as \emph{speaker diarization}. Speaker diarization has been studied extensively. Miro et al.~\cite{miro12} offer a comprehensive review of recent research in this field. In realistic situations, with far-field microphones, or microphone arrays, the task of active speaker detection from audio is far from trivial. Most methods (e.g.,~\cite{anguera06,fredouille06}), use some form of model-based supervised training. This is one of the motivation for our study: firstly, we believe that complementing the auditory modality with visual information can be useful if not necessary for this task, especially in the more challenging acoustic conditions. Secondly, we want to comply with a developmental approach, where the learning system only uses the information available through its senses in the interaction with humans. We therefore want to avoid the need for careful annotations that are required by the aforementioned supervised methods.

Audio-visual speaker detection combines information from both the audio and the video signals. The application of audio-visual synchronization to speaker detection in broadcast videos was explored by Nock et al.~\cite{nock03}. Unsupervised audio-visual detection of the speaker in meetings was proposed in~\cite{friedland09}. Zhang et al.~\cite{zhang08} presented a boosting-based multi-modal speaker detection algorithm applied to distributed meetings, to give three examples. Mutual correlations to associate an audio source with regions in the video signal was demonstrated by Fisher et al.~\cite{fisher00}, and Slaney and Covell~\cite{slaney00} showed that audio-visual correlation can be used to find the temporal synchronization between audio signal and a speaking face. An elegant solution was proposed in~\cite{hershey00} where the mutual information between the acoustic and visual signals is computed by means of a joint multivariate Gaussian process, with the assumption that only one audio and one video streams were present and that locating the source corresponds to finding the pixels in the image that correlate with acoustic activity. In more recent studies, researchers have employed artificial neural network architectures to build active speaker detectors from audio-visual input. A multi-modal Long Short-Term Memory model that learns shared weights between modalities was proposed in~\cite{ren16}. The model was applied to speaker naming in TV shows. Hu et al.~\cite{hu15} proposed a Convolutional Neural Network model that learns the fusion function of face and audio information.

Other approaches for speaker detection include a general pattern recognition framework used by Besson and Kunt~\cite{besson08} applied to detection of the speaker in audio-visual sequences. Visual activity (the amount of movement) and focus of visual attention were used as inputs by Hung and Ba~\cite{hung09} to determine the current speaker on real meetings. Stefanov et al.~\cite{stefanov16_2} used action units as inputs to Hidden Markov Models to determine the active speaker in multi-party interactions and Vajaria et al.~\cite{vajaria08} demonstrated that information for body movements can improve the detection performance.

Most of the approaches cited in this section are either evaluated on small amounts of data, or have not been proved to be usable in real-time settings. Furthermore, they usually require manual annotations and the spatial configuration of the interaction and the relative position of the input sensors is known. The goal is usually an offline video/audio analysis task, such as semantic indexing and retrieval of TV broadcasts or meetings, or video/audio summarization. We believe that the challenge of real-time detection of the active speaker in dynamic and cluttered environments remains. In the context of automatic language acquisition, we want to infer the possible objects the active speaker is focusing attention on. In this context, assumptions such as known sensor arrangement or participants' position and number in the environment are unrealistic, and should be avoided. Therefore, in this study we present methods which have several desirable characteristics for such types of scenarios: 1) they work in real-time, 2) they do not assume specific spatial configuration (sensors or participants), 3) the number of possible (simultaneously) speaking participants is free to change during the interaction, and 4) no externally produced labels are required, but rather the acoustic inputs are used as reference to the visually based learning.

\section{Methods}
\label{sec:methods}
The goal of the methods described in this section is to detect in real-time the state (speaking or not speaking) of all visible faces in a multi-person language learning scenario, using only visual information (the RGB color data). An illustration of the desired output of an active speaker detector can be seen in Figure~\ref{fig:detection_example}.

We use a self-supervised learning approach to construct an active speaker detector: the machine learning methods are supervised, but the labels are obtained automatically from the auditory modality to learn models in the visual modality. An overview of the approaches considered in the study is given in Figure~\ref{fig:methods}. The first row in the figure illustrates the perceptual inputs that are automatically extracted from the raw audio and video streams. The visual input consists of RGB images of each face extracted from the video stream with the Viola and Jones's face detector~\cite{viola01}. The auditory input consists of labels extracted from the audio stream which correspond to the voice activity. The used audio-only voice activity detector (VAD)~\cite{skantze12} is based on two thresholds on the energy of the signal, one to start a speech segment and one to end it. These thresholds are adaptive and based on a histogram method. The ability to extract face images and VAD labels is given as a starting point to the system and is motivated in Section~\ref{sec:discussion}.

The methods use a feature extractor based on a Convolutional Neural Network, followed by a classifier. Two types of classifiers are tested: non-temporal (Perceptron) and temporal (Long Short-Term Memory Network). Additionally, two techniques for training the models are considered: transfer learning that employs a pre-trained feature extractor and only trains a classifier specifically for the task; and task specific learning that trains a feature extractor and a classifier simultaneously for the task.

Each method outputs a posterior probability distribution over the two possible outcomes (speaking or not speaking). Since the goal is a binary classification, the detection of the active speaker happens when the corresponding probability exceeds $0.5$. The evaluation of each method is performed by computing the accuracy of the predictions on frame-by-frame basis (Section~\ref{sec:experiments}).

\subsection{Task Specific Learning}
\label{sec:task_specific_learning}
An illustration of the task specific learning method is shown in the second row of Figure~\ref{fig:methods}. This method trains a Convolutional Neural Network (CNN) feature extractor in combination with a Perceptron classifier with the goal of classifying each input image either as speaking or not speaking. During the training phase both images and labels are used by a gradient-based optimization procedure~\cite{kingma14} to adjust the weights of the CNN and Perceptron models. During the prediction phase, only images are used by the trained models to generate labels. The CNN and Perceptron models work on a frame-by-frame basis and have no memory of past frames.

\subsection{Transfer Learning}
\label{sec:transfer_learning}
An illustration of this method can be seen in the third row of Figure~\ref{fig:methods}.
Similarly to the previous method, the transfer learning method uses a CNN and a Perceptron model. In this method, however, the CNN model is pre-trained on an object recognition task (i.e., VGG16~\cite{simonyan14}). To adapt the VGG16 model to the active speaker detection task, the object classification layer is removed and the truncated VGG16 model is used as a feature extractor. Then the method consists of training only a Perceptron model to map the features generated by the VGG16 model to the speaker activity information. As for the task specific learning method, this method has no memory of past frames.

Because the VGG16 model was originally trained in a supervised manner to classify objects, this raises the question on how suitable this model is in the context of developmental language acquisition. Support to the use of this model comes from the literature  on visual perception that demonstrates the ability of infants to recognize objects very early in their development~\cite{spelke90,kirkham02}.

\subsection{Temporal Learning}
\label{sec:temporal_learning}
The temporal learning method is illustrated by the forth row of Figure~\ref{fig:methods}. This method is based on the previously described feature extractors, but introduces a model of the time evolution of the perceptual inputs. During the training phase a custom (CNN) or pre-trained (VGG16) feature extractor constructs a feature vector for each input image. Then the features and labels are used by a gradient-based optimization procedure~\cite{kingma14} to adjust the weights of a Long Short-Term Memory (LSTM) model~\cite{hochreiter97}. During the prediction phase, images are converted into features with a custom CNN or VGG16 model, which features are then used by the trained detector (LSTM) to generate labels.

\subsection{Acoustic Noise}
\label{sec:acoustic_noise}
In order to test the effect of noise on the audio-only VAD, stationary noise is added to the audio signal. The noise is sampled from a Gaussian distribution with zero mean and variance $\sigma^2$. For every recording, the active segments are first located by means of the audio-only VAD. These are then used to estimate the energy $E_x$ of the signal as the mean squares of the samples. Then $\sigma^2$ is computed as the ratio between the energy of the signal and the desired signal-to-noise ratio (SNR):

\begin{equation}
\sigma^2 = \frac{E_x}{10^{\frac{\text{SNR}}{10}}}.
\end{equation}
Finally, the noise is added to the signal, and the samples are re-normalized to fit in the 16 bit linear representation. The audio-only VAD is used again on the noisy signal and its accuracy is computed on the result.

\begin{figure}[!t]
\centering
\includegraphics[width=0.75\columnwidth]{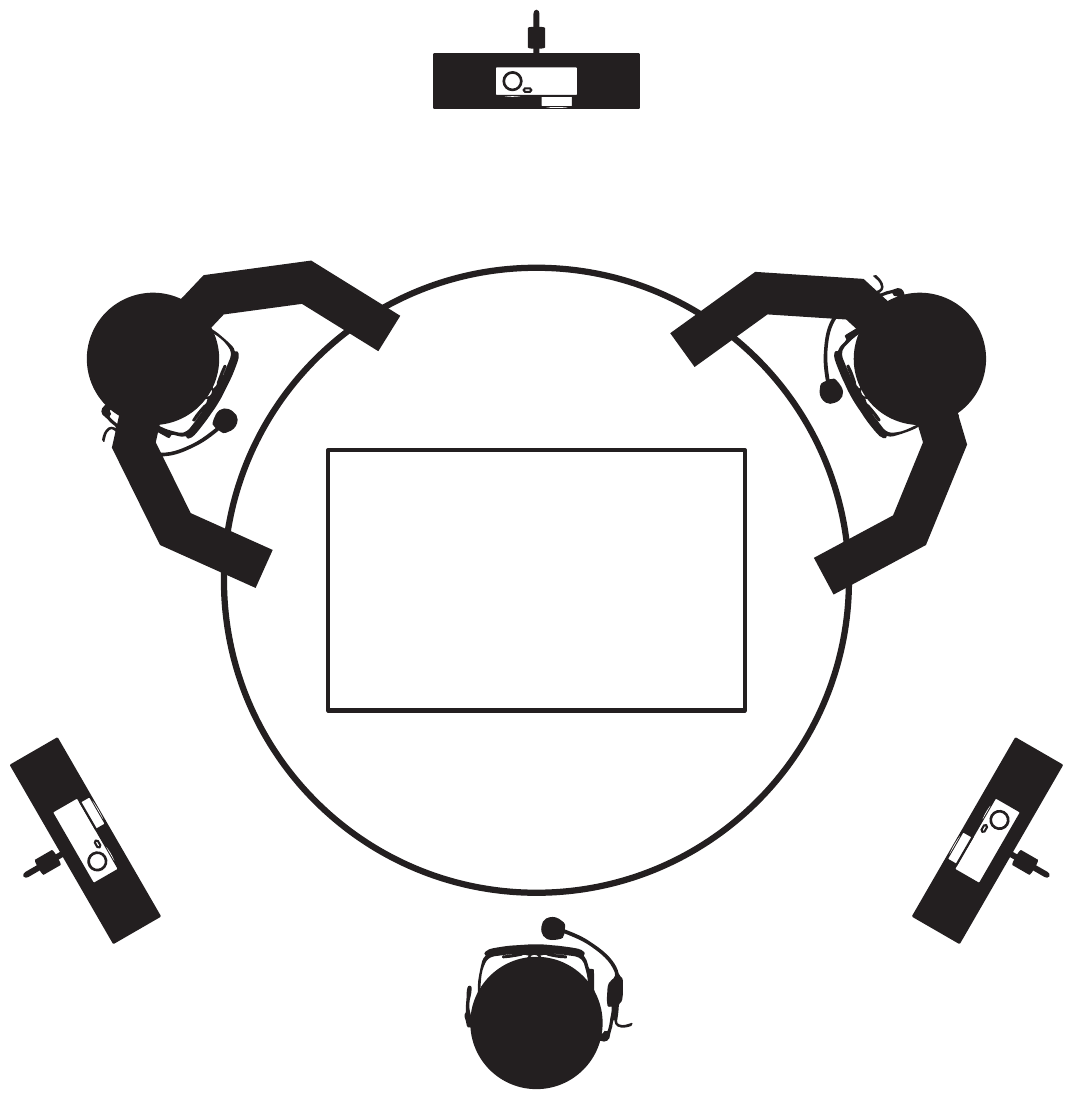}
\caption{Spatial configuration of the sensors and participants in the dataset.}
\label{fig:dataset_setup}
\end{figure}

\section{Experiments}
\label{sec:experiments}
This section is divided in two parts. The first part describes the dataset used to build and evaluate the active speaker detectors. The second part describes the general setup of the conducted experiments.

\subsection{Dataset}
\label{sec:dataset}
The methods presented in Section~\ref{sec:methods} are implemented and evaluated using a multimodal multiparty interaction dataset described in~\cite{stefanov16_1}. The main purpose of the dataset is to explore patterns in the focus of visual attention of humans under the following three different conditions: two humans involved in task-based interaction with a robot; the same two humans involved in task-based interaction where the robot is replaced by a third human, and a free three-party human interaction. The dataset contains two parts: 6 sessions with duration of approximately 30 minutes each, and 9 sessions, each of which is with duration of approximately 40 minutes. The dataset is rich in modalities and recorded data streams. It includes the streams generated from 3 Kinect v2 devices (color, depth, infrared, body and face data), 3 high quality audio streams generated from close-talking microphones, 3 high resolution video streams generated from GoPro cameras, touch-events stream for the task-based interactions generated from an interactive surface, and the system state stream generated by the robot involved in the first condition. The second part of the dataset also includes the data streams generated from 3 Tobii Pro Glasses 2 eye trackers. The interactions are in English and all data streams are spatially and temporally synchronized and aligned. The interactions occur around a round interactive surface and all 24 unique participants are seated. Figure~\ref{fig:dataset_setup} illustrates the spatial configuration of the setup in the dataset.

\begin{figure}[!t]
\centering
\includegraphics[width=\columnwidth]{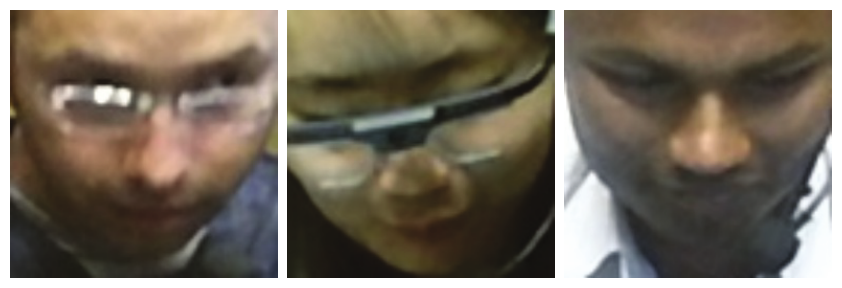}
\caption{Example of a difficult visual input from the first and second condition in the dataset.}
\label{fig:dataset_example}
\end{figure}

As described previously, each interaction in the dataset is divided into three conditions, with the first and second condition being related to a collaborative task-based interaction in which the participants play a game on a touch surface. During this two conditions the participants interact mainly with the touch surface and discuss with their partner how to solve the given task. Therefore, the participants' overall gaze direction (head orientation) is towards the touch surface. This raises some very challenging visual conditions for extracting speech activity information from the face. We show three examples in Figure~\ref{fig:dataset_example}. This observation motivated experiments using only the data from the third condition of each interaction.

\begin{figure*}[!t]
\centering
\includegraphics[width=\textwidth]{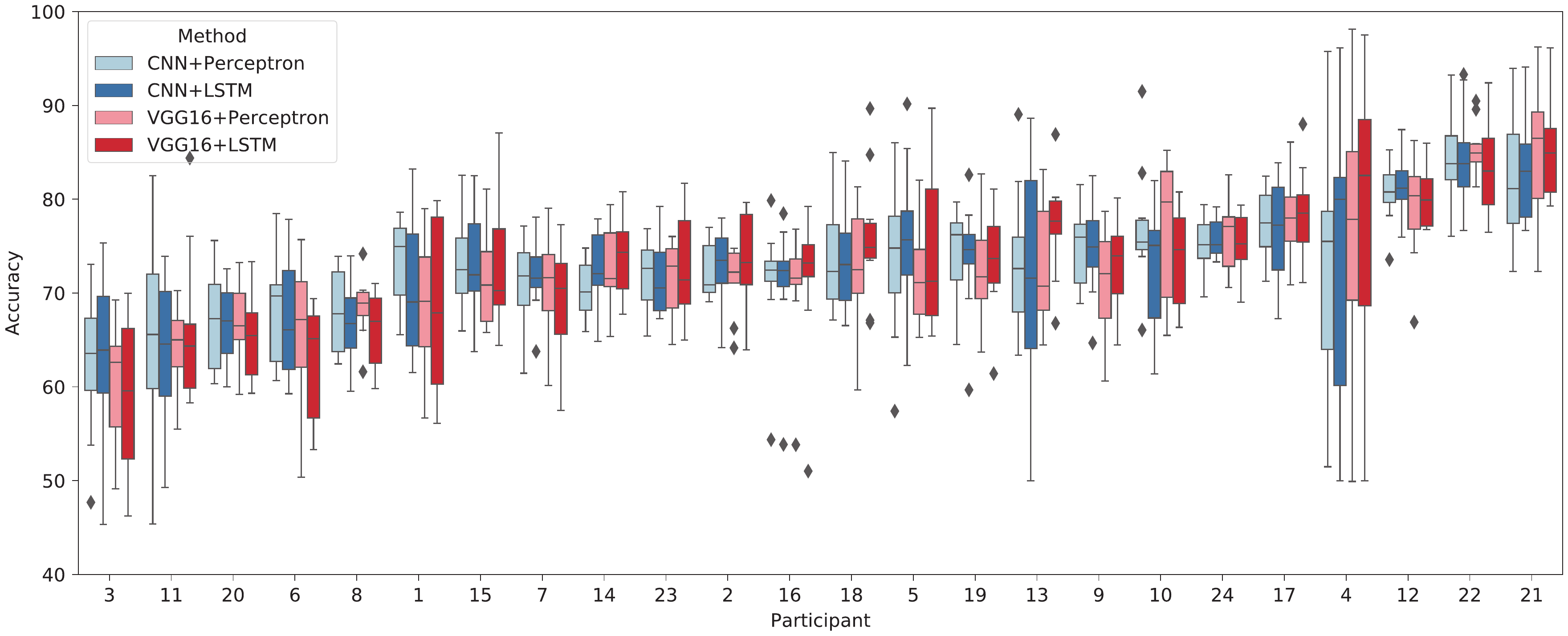}
\caption{Accuracy versus participant and method. The participants are sorted by overall accuracy. The segment length for the LSTMs is 15 frames (500 ms). The boxplots show the results over all 10 folds.}
\label{fig:sd_results}
\end{figure*}

\begin{table*}[!t]
\caption{Speaker dependent results (10-fold cross-validation); Mean accuracy and standard deviation.}
\label{tab:sd_results}
\centering
\begin{tabular}{ c | c c c c c }
\textbf{Features} & \textbf{Perceptron} & \textbf{LSTM\_15} & \textbf{LSTM\_30} & \textbf{LSTM\_150} & \textbf{LSTM\_300} \\
\hline
CNN & \textbf{73.13} (7.81) & 72.92 (8.47) & \textbf{73.13} (8.67) & 72.61 (9.54) & 72.46 (9.56) \\
VGG16 & 72.61 (8.27) & 72.90 (8.85) & \textbf{73.27} (9.14) & 72.46 (9.97) & 72.55 (10.22)
\end{tabular}
\end{table*}

\subsection{Experimental Setup}
\label{sec:experimental_setup}
This section describes the general setup of the experiments. In all experiments the video stream is generated by the Kinect v2 device directed at the participant under consideration and the audio stream is generated by the participant's close-talking microphone. The total amount of frames used in the experiments is 690000 (${\sim}6.5$ hours).

The CNN models comprise three convolutional layers of width 32, 32, and 64 with receptive fields of $3\times3$ and rectifier activations, interleaved by max pooling layers with window size of $2\times2$. The output of the last max pooling layer is used by a densely connected layer of size 64 with rectifier activation functions and finally by a perceptron layer with logistic sigmoid activations. The LSTM models include one long short-term memory layer of size 128 with hyperbolic tangent activations, followed by a densely connected and a perceptron layer similarly to the CNN models.

During the training phase the models use Adam optimizer with default parameters ($\alpha=0.001$, $\beta_{1}=0.9$, $\beta_{2}=0.999$, and $\epsilon=10^{-8}$) and binary crossentropy loss function. Each non-temporal model (CNN and Perceptron) is trained for $50$ epochs and each temporal model (LSTM) is trained for $100$ epochs. The LSTM models are trained with 15, 30, 150, and 300 frame (500 ms, 1 s, 5 s, and 10 s) long segments without overlaps. The models corresponding to the best validation performance are selected for evaluation on the test set. The models are implemented in Keras~\cite{chollet15} with TensorFlow~\cite{tensorflow15} backend. During the prediction phase only the RGB color images extracted with the face detector are used as input. As described previously, each of the considered methods outputs a posterior probability distribution over the two possible outcomes --- speaking or not speaking. Therefore, when evaluating the models' performance, 0.5 is used as a threshold for assigning a class to each frame-level prediction. The results are reported in terms of frame-by-frame weighted accuracy which is calculated with,

\newcommand{\acc}{\ensuremath \text{wacc}}
\newcommand{\tp}{\ensuremath \text{tp}}
\newcommand{\fp}{\ensuremath \text{fp}}
\newcommand{\tn}{\ensuremath \text{tn}}
\newcommand{\fn}{\ensuremath \text{fn}}
\begin{equation}
\acc = 100 \times \frac{\frac{\tp}{\tp+\fn} + \frac{\tn}{\fp+\tn}}{2},
\end{equation}
where $\tp$, $\fp$, $\tn$, and $\fn$ are the number of true positives, false positives, true negatives and false negatives, respectively. As a consequence, regardless of the actual class distribution in the test set (which is in general different for each participant), the baseline chance performance using this metric is always 50\%. Although this metric allows an easy comparison of results between different participants and methods, it is a very conservative measure of performance (Section~\ref{sec:metric}).

The study presents three experiments with the proposed methods: speaker dependent, multi-speaker dependent, and speaker independent. The speaker dependent experiment builds a model for each participant and tests it on independent data from the same participant. This process is repeated 10 times per participant with splits generated through a 10-fold cross-validation procedure. The multi-speaker dependent experiment uses the splits generated in speaker dependent experiment. This experiment, however, builds a model with the data for all participants and tests it on the independent data from all participants. This experiment tests the scalability of the proposed methods to more than one participant. The speaker independent experiment uses a leave-one-out cross-validation procedure to build and evaluate the models. This experiment tests the transferability of the proposed methods to unseen participants.

Finally, as described in Section~\ref{sec:acoustic_noise}, the effect of noise is tested on the audio-only VAD. The proposed video-only active speaker detectors are compared with audio-only VAD where the SNR varies from 0 to 30 in increments of 5.

\begin{figure*}[!t]
\centering
\includegraphics[width=0.32\textwidth]{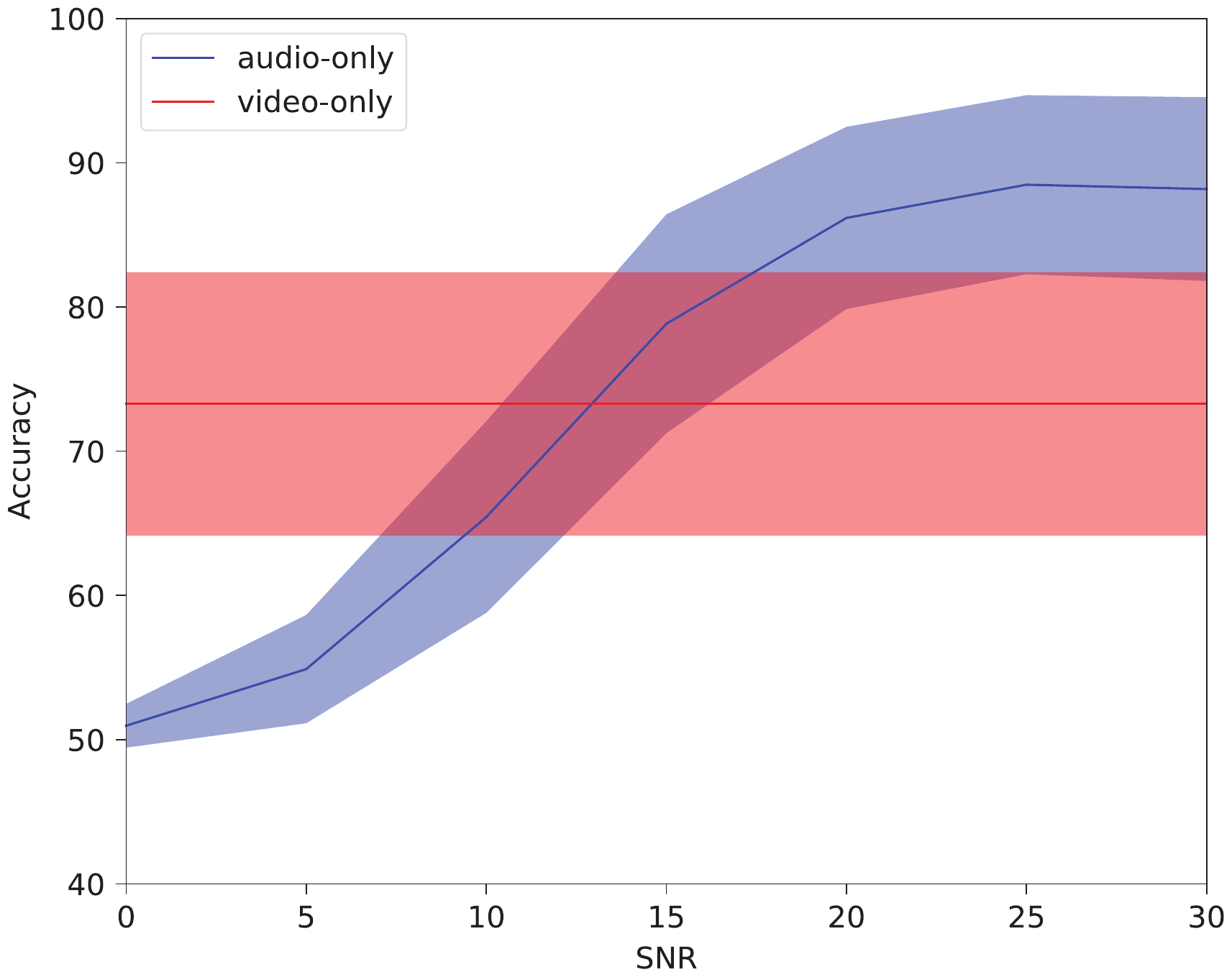}
\includegraphics[width=0.32\textwidth]{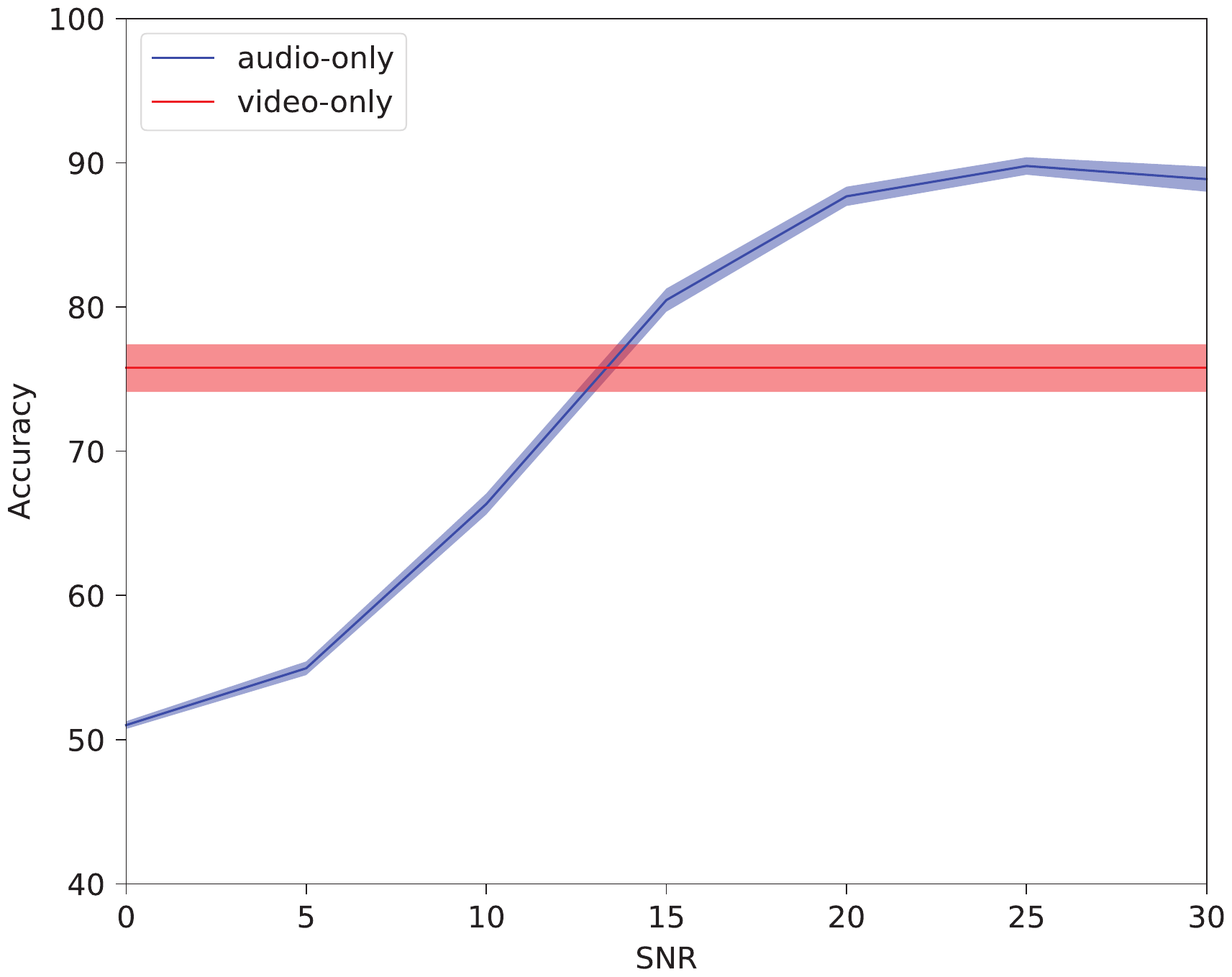}
\includegraphics[width=0.32\textwidth]{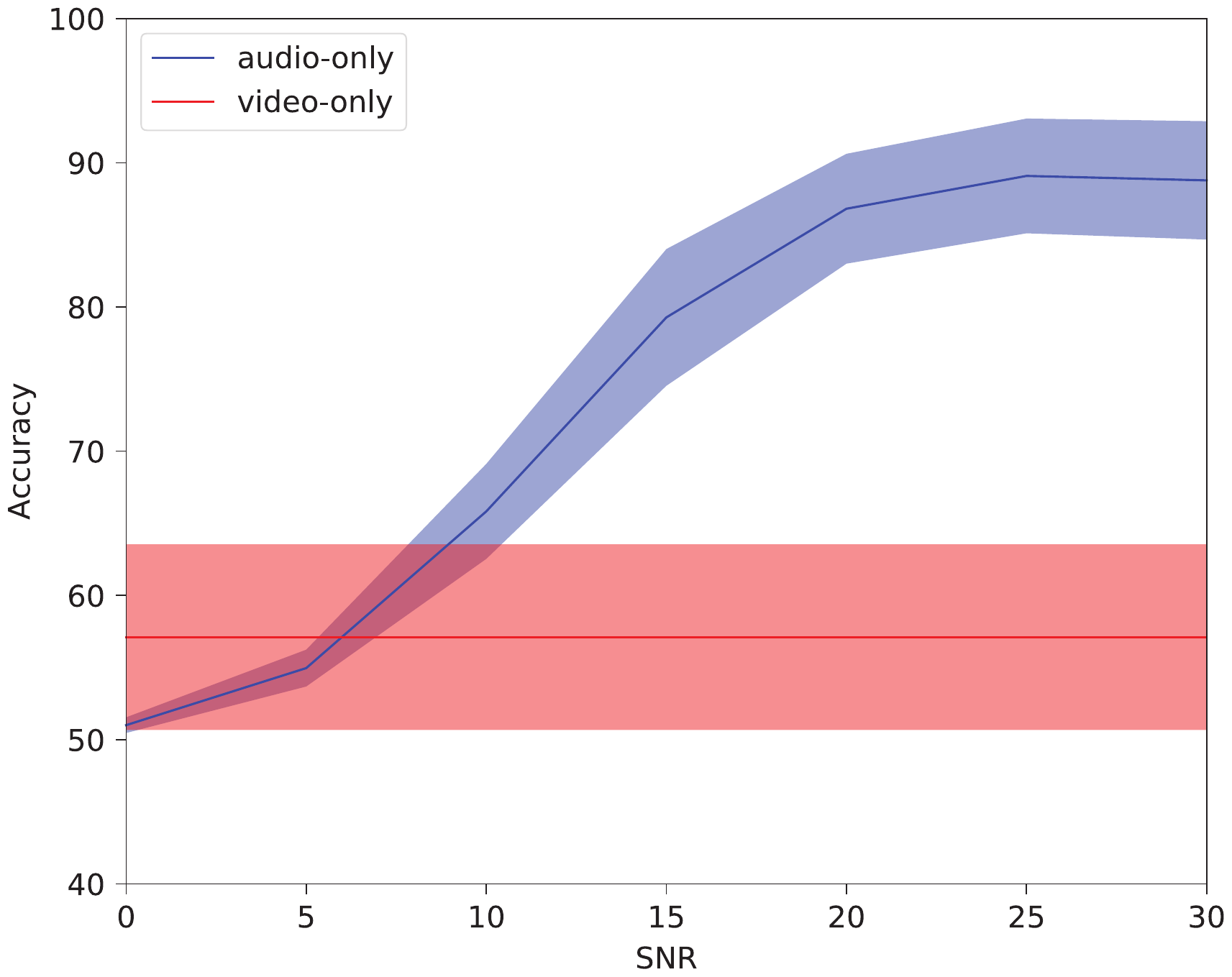}
\caption{Comparison between audio-only and video-only method in noise (the solid lines are accuracies and the shaded areas are standard deviations). The accuracy in the speaker dependent experiment (left) is averaged over 24 participants and 10 folds. The accuracy in the multi-speaker dependent experiment (center) is averaged over 10 folds each containing data from 24 participants. The accuracy in the speaker independent experiment (right) is averaged over 24 folds each containing data from the participant that was left out during training. In all cases the performance  of the audio-only method degrades with the reduction in SNR, whereas the video-only method is not affected by acoustic noise.}
\label{fig:comparison}
\end{figure*}

\section{Results}
\label{sec:results}
This section presents the numerical results obtained from the experiments.

\subsection{Speaker Dependent}
\label{subsec:speaker_dependent}
The mean accuracy and standard deviation per method obtained in the speaker dependent experiment are provided in Table~\ref{tab:sd_results}. The highest mean result in this experiment is 73.13\% for the LSTM\_30 models when using custom CNN feature extractors and 73.27\% for the LSTM\_30 models when using pre-trained VGG16 feature extractors. The complete results are illustrated in Figure~\ref{fig:sd_results}. The figure shows that the accuracy varies significantly between participants. Also the variability between participants is higher than the difference obtained with different methods per participant. A comparison between the best performing video-only method and an audio-only VAD is illustrated in the left plot of Figure~\ref{fig:comparison}. The two methods give similar results for a range of SNRs around 12. The video-only method outperforms the audio-only VAD for more noisy conditions, whereas the opposite is true if the SNR is greater than 20.

\subsection{Multi-Speaker Dependent}
\label{subsec:multi-speaker_dependent}
The summarized results of the multi-speaker dependent experiment are provided in Table~\ref{tab:msd_results}. The highest mean result in this experiment is 75.76\% for the LSTM\_150 models when using custom CNN feature extractors. A comparison between the best performing video-only method and an audio-only VAD is illustrated in the center plot of Figure~\ref{fig:comparison}. Similarly to the speaker dependent case, the two methods give similar results for a range of SNRs around 12. However, in this case the spread around the mean is much reduced because every fold includes a large collection of samples from all participants.

\begin{table*}[!t]
\caption{Multi-speaker dependent results (10-fold cross-validation); Mean accuracy and standard deviation.}
\label{tab:msd_results}
\centering
\begin{tabular}{ c | c c c c c }
\textbf{Features} & \textbf{Perceptron} & \textbf{LSTM\_15} & \textbf{LSTM\_30} & \textbf{LSTM\_150} & \textbf{LSTM\_300} \\
\hline
CNN & 74.80 (1.63) & 74.91  (1.54) & 75.11 (1.57) & \textbf{75.76} (1.65) & 75.26 (1.46)
\end{tabular}
\end{table*}

\subsection{Speaker Independent}
\label{subsec:speaker_independent}
The summarized results of the speaker independent experiment are provided in Table~\ref{tab:si_results}. The highest mean result in this experiment is 57.11\% for the LSTM\_30 models when using custom CNN feature extractors. A comparison between the best performing video-only method and an audio-only VAD is illustrated in the right plot of Figure~\ref{fig:comparison}. As can be observed, the results from the video-only method are only slightly above chance level, hence falling far behind the audio-based VAD.

\begin{table*}[!t]
\caption{Speaker Independent Results (leave-one-out cross-validation); Mean accuracy and standard deviation.}
\label{tab:si_results}
\centering
\begin{tabular}{ c | c c c c c }
\textbf{Features} & \textbf{Perceptron} & \textbf{LSTM\_15} & \textbf{LSTM\_30} & \textbf{LSTM\_150} & \textbf{LSTM\_300} \\
\hline
CNN & 55.39 (5.74) & 56.33 (6.56) & \textbf{57.11} (6.44) & 56.96 (6.50) & 57.55 (7.02)
\end{tabular}
\end{table*}

\section{Discussion}
\label{sec:discussion}
In order to interpret the results presented in Section~\ref{sec:results} we need to make a number of considerations about the evaluation method. We will also consider the advantages and limitations of the metric used and detail the assumptions made in the methods and the main contributions of the study.

The proposed methods estimate the probability of speaking independently for each face. This has the advantage of being able to detect several speakers that are active at the same time, but for many applications it might be sufficient to select the active speaker among the detected faces. Doing this would allow us to combine the single predictions into a joint probability, thus increasing the performance.

It is important to note that the conditions in the experiment that compared audio-only and video-only methods were favorable to the audio-only method due to the use of stationary noise. The VAD employed for the audio-based detection uses adaptive thresholds that are specifically suitable for stationary noise. Therefore we would expect a larger advantage for the video-based speaker detection in low to medium SNRs in the presence of non-stationary noises often present in natural communication environments.

\subsection{Metric}
\label{sec:metric}
Evaluating the proposed methods on a frame-by-frame basis gives a detailed measure of performance. However, one might argue that frame-level (33 ms) accuracy is not necessary for artificial cognitive systems employing the proposed methods in the context of automatic language acquisition. Evaluating the methods on a fixed-length sliding time window (e.g., 200 ms) might be sufficient for this application.

Furthermore, the definition of the weighted accuracy amplifies short mistakes. For example, if in 100 frames, 98 belong to the active class and 2 to the inactive class, a method that classifies all frames as active will have $\acc = \frac{100}{2} \times \left[\frac{98}{98+0} + \frac{0}{2+0}\right] = 50\%$. If we consider a case of continuous talking, where the speaker takes short pauses to recollect a memory or structure the argument, then a perfect audio-only method will detect silence of certain length (at least 200 ms) in the acoustic signal and label the corresponding video frames as not speaking. However, from the interaction point of view the speaker might be still active, resulting also in visual activity. A video-only method that misses these short pauses would be strongly penalized by the used metric, achieving as low as 50\% accuracy when all other frames are classified correctly. Similar situation occurs when a person is listening and gives short acoustic feedbacks which are missed by the video-only methods.

The advantage of the weighted accuracy metric, however, is that it enables us to seamlessly compare the performance between participants and methods. This is because, the different underlying class distributions due to each particular dataset, are accounted for by the metric and the resulting baseline is 50\% for all considered experimental configurations.

\subsection{Assumptions}
\label{sec:assumptions}
The proposed methods make the following assumptions:
\begin{itemize}
\item the system is able to detect faces,
\item the system is able to detect speech for a single speaker,
\item there are situations in which the system only interacts with one speaker, and can therefore use the audio-only VAD to train the video-only active speaker detector.
\end{itemize}
In order to motivate the plausibility of these assumptions in the context of a computational method for language acquisition, we consider research in developmental psychology. According to studies reported in~\cite{labarbera76, browne77} infants can discriminate between several facial expressions which suggests that they are capable of detecting human faces. The assumption that the system can detect speech seems to be supported by research on recognition of mother's voice in infants (e.g.,~\cite{mills74}). However, whereas infants can detect the voice at a certain distance from the speaker, here we make the simplifying assumption that we can record and detect speech activity from close-talking microphones for each speaker. It remains to be verified if we can obtain similar performance from the audio-only VAD in case we use far-field microphones or microphone arrays, or in noisy acoustic conditions. The final assumption is reasonable considering that infants interact with small number of speakers in their first months, and in many cases only one parent is available as caregiver at any specific time. 

\subsection{Contributions}
\label{sec:contributions}
This study extends our previous work~\cite{stefanov17} on vision-based methods for detection of the active speaker in multi-party human-robot interactions. We will summarize the main differences between this study and~\cite{stefanov17} in this section. The first difference is the use of a better performing pre-trained CNN model for feature extraction (i.e., VGG16~\cite{simonyan14}) compared to the previously used AlexNet~\cite{krizhevsky12}. We also significantly extended the set of experiments to evaluate and compare the proposed methods. In the current study we evaluated the effect of using temporal models by comparing the performance of LSTM models similar to the ones evaluated in~\cite{stefanov17}, to non-temporal Perceptron models. Furthermore, we compared the performance of transfer learning models, with models that are built specifically for the current application and trained exclusively on the task specific data. Finally we reported results for multi-speaker and speaker independent experiments.

One of our findings is that, given that we optimize the classifier to the task (Perceptron or LSTM), it is not necessary to optimize the feature extractor (the custom CNNs perform similarly to the pre-trained VGG16). This suggests that a pre-trained feature extractor such as VGG16 works well independently of the speaker and can be used to extend the results beyond the participants in the present dataset. Also, the result of the multi-speaker dependent experiment shows that the proposed methods can scale beyond a single speaker without decrease in performance. Combining this observation with the observation for the applicability of transfer learning suggests that a mixture of the proposed methods can be indeed an useful component of a real life artificial cognitive system.

Finally, the speaker independent experiment yields significantly lower performance compared to the other two experiments. We should mention, however, that, from a cognitive system's perspective, this might be an unnecessarily challenging condition. We can in fact expect infants to be familiar with a number of caregivers, thus justifying a condition more similar to the settings in the multi-speaker dependent experiment.

\section{Conclusions}
\label{sec:conclusions}
In this study we proposed and evaluated three methods for automatic detection of the active speaker based solely on visual input. The proposed methods are intended to complement acoustic methods, especially in noisy conditions, and could assist an artificial cognitive system to engage in social interactions which has been shown to be beneficial for language acquisition.

We tried to reduce the assumptions about the language learning environment to a minimum. Therefore, the proposed methods allow different speakers to speak simultaneously as well as to be all silent; the methods do not assume a specific number of speakers, and the probability of speaking is estimated independently for each speakers, thus allowing the number of speakers to change during the social interaction.

We evaluated the proposed methods on a large multi-person dataset. The methods perform well on a speaker dependent and multi-speaker dependent fashion, reaching accuracy of over 75\% (baseline 50\%) on a weighted frame-based evaluation metric. The combined results obtained from the transfer learning and multi-speaker learning experiments are promising and suggest that the proposed methods can generalize to unseen perceptual inputs by incorporating a model adaptation step for each new face.

We should acknowledge the general difficulty of the problem addressed in this study. Humans generally produce many facial configurations when they are not speaking that might be highly overlapping to the configurations associated with when they are speaking.

The methods proposed in this study are in support to socially-aware language acquisition and they can be seen as mechanisms for constraining the visual input thus providing higher quality and more appropriate data for a statistical learning of word-referent mappings. Therefore, the main purpose of the methods is to help bringing an artificial cognitive system one step closer to resolving the referential ambiguity in cluttered, dynamic, and noisy environments.

\IEEEtriggeratref{48}
\bibliographystyle{IEEEtran}
\bibliography{IEEEabrv,article}

\begin{IEEEbiography}[{\includegraphics[width=1in,keepaspectratio]{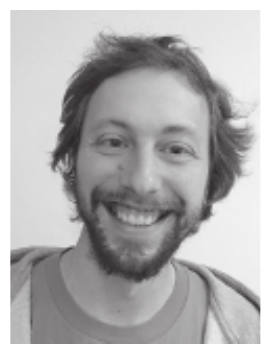}}]{Kalin Stefanov}
received the MSc degree in Artificial Intelligence from University of Amsterdam (Amsterdam, The Netherlands) and the PhD degree in Computer Science from KTH Royal Institute of Technology (Stockholm, Sweden). He is currently a post-doctoral fellow at the Institute for Creative Technologies, University of Southern California (Los Angeles, USA). His research interests include machine learning, computer vision and speech technology.
\end{IEEEbiography}

\begin{IEEEbiography}[{\includegraphics[width=1in,keepaspectratio]{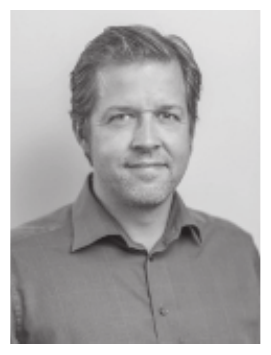}}]{Jonas Beskow}
is a Professor in Speech Communication with research interests in multimodal speech technology, modeling and generating verbal and non-verbal communicative behavior as well as embodied conversational agents or social robots that use speech, gesture and/or other modalities in order to accomplish human-like interaction. He is also a co-founder of Furhat Robotics, a startup developing an innovative social robotics platform based on KTH research.
\end{IEEEbiography}

\begin{IEEEbiography}[{\includegraphics[width=1in,keepaspectratio]{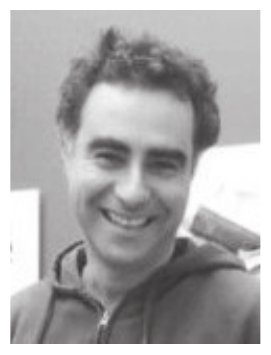}}]{Giampiero Salvi}
received the MSc degree in Electrical Engineering from Università la Sapienza (Rome, Italy) and the PhD degree in Computer Science from KTH Royal Institute of Technology (Stockholm, Sweden). He was a post-doctoral fellow at the Institute of Systems and Robotics (Lisbon, Portugal). He is currently Professor in Machine Learning at NTNU Norwegian University of Science and Technology (Trondheim, Norway). His main interests are machine learning, speech technology, and cognitive systems.
\end{IEEEbiography}
\vfill

\end{document}